\documentclass[10pt,twocolumn,letterpaper]{article}

\usepackage{iccv}
\usepackage{times}
\usepackage{epsfig}
\usepackage{graphicx}
\usepackage{amsmath}
\usepackage{amssymb}

\usepackage{booktabs}
\usepackage{enumitem}
\usepackage{breakcites}

\usepackage{tikz}
\usepackage{comment}
\usepackage{amsmath,amssymb} % define this before the line numbering.
\usepackage{color}
\usepackage{booktabs}
\usepackage{multirow}
\usepackage{graphicx, wrapfig, lipsum}
\usepackage{sidecap}
\usepackage{soul}

\usepackage[pagebackref,breaklinks,colorlinks]{hyperref}

\usepackage[accsupp]{axessibility}  % Improves PDF readability for those with disabilities.

\usepackage[capitalize]{cleveref}
\crefname{section}{Sec.}{Secs.}
\Crefname{section}{Section}{Sections}
\Crefname{table}{Table}{Tables}
\crefname{table}{Tab.}{Tabs.}

\usepackage{xcolor}

\iccvfinalcopy % *** Uncomment this line for the final submission

\ificcvfinal\fi

\begin{document}

\title{MSI: Maximize Support-Set Information for Few-Shot Segmentation}

\author{Seonghyeon Moon\\
Rutgers University\\
{\tt\small sm2062@rutgers.edu}
\and
Samuel S. Sohn\\
Rutgers University\\
{\tt\small samuel.sohn@rutgers.edu}
%
% \and
% Honglu Zhou\\
% Rutgers University\\
% {\tt\small honglu.zhou@rutgers.edu}
\and
Honglu Zhou\\
NEC Laboratories America\\
{\tt\small hozhou@nec-labs.com}
\and
Sejong Yoon\\
The College of New Jersey\\
{\tt\small yoons@tcnj.edu}
\and
Vladimir Pavlovic\\
Rutgers University\\
{\tt\small vladimir@rutgers.edu}
\and
Muhammad Haris Khan\\
Mohamed Bin Zayed University of Artificial Intelligence\\
{\tt\small muhammad.haris@mbzuai.ac.ae}
\and
Mubbasir Kapadia\\
Rutgers University\\
{\tt\small mubbasir.kapadia@rutgers.edu}
}

\maketitle
\ificcvfinal\thispagestyle{empty}\fi

%%%%%%%%% ABSTRACT
\begin{abstract}
FSS~(Few-shot segmentation)~aims to segment a target class using a small number of labeled images (support set). To extract information relevant to the target class, a dominant approach in best performing FSS methods
% baselines 
removes background features using a support mask. We observe that this feature excision through a limiting support mask introduces an information bottleneck in several challenging FSS cases, e.g., for small targets and/or inaccurate target boundaries. To this end, we present a novel method~(MSI), which maximizes the support-set information by exploiting two complementary sources of features to generate super correlation maps. We validate the effectiveness of our approach by instantiating it into three recent and strong FSS methods.
% baselines. 
Experimental results on several publicly available FSS benchmarks show that our proposed method consistently improves performance by visible margins and leads to faster convergence. Our code and trained models are available at: \url{https://github.com/moonsh/MSI-Maximize-Support-Set-Information}
\end{abstract}

%%%%%%%%% BODY TEXT
%\vspace{-0.2in}
\section{Introduction}
\label{sec:intro}

\begin{figure}[t]
  \centering
    \includegraphics[height=6.6cm]{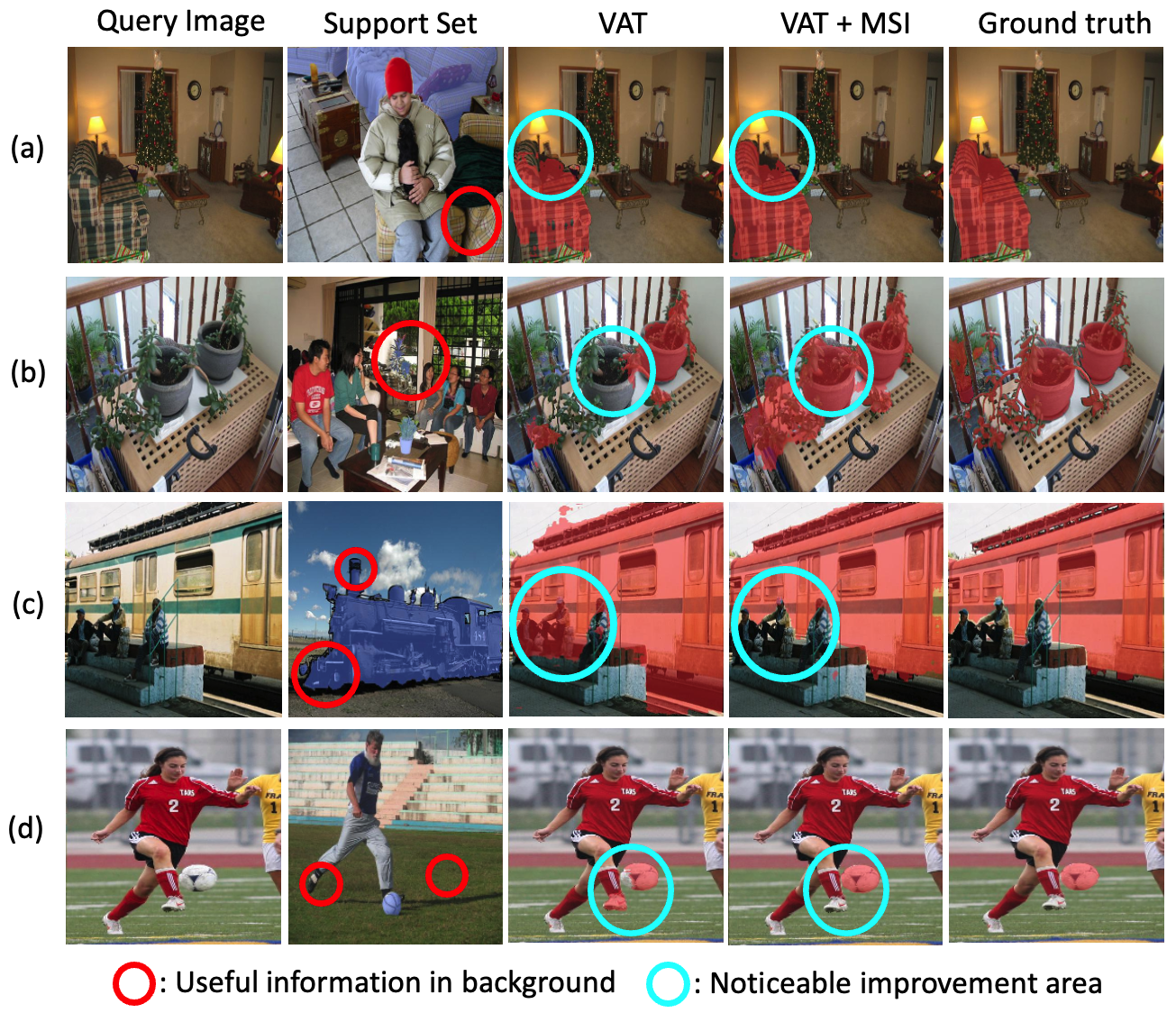}

   \caption{Recent FSS baseline (VAT\cite{VAT2}) struggles to accurately segment the target object in several challenging scenarios in PASCAL~\cite{pascal} and COCO$^i$~\cite{lin2015microsoft}: (a) the same instance of the target class is not masked, e.g., the sofa on the right, (b) the support mask is very small compared to the entire image, e.g., flowers in the pot, (c) support mask is missing some target boundary information, e.g., the front and chimney of the train, and (d) the background contains some important contextual information, unavailable in the support mask, e.g., shoes and grass. Our method (MSI) is capable of accurately segmenting target objects. It maximizes the support set information to compensate for the limited support mask information and can exploit relevant contextual information from the background.} 
   
%   is capable of accurately segmenting target objects when there is limited information in the support mask and some context information is
   
%   Cases beneficial to utilize background in support image from PASCAL~\cite{pascal} and COCO$^i$~\cite{lin2015microsoft} where proposed method MSI~(Ours) performs accurate segmentation compared to VAT~\cite{VAT} with ResNet50~\cite{Resnet}. 
   %\vspace{-0.2in}
%   (a)~The same target class, the sofa on the right side, is not masked. (b)~The support mask is very small therefore removing the background results in leaving very limited information. (c)~Support mask is not perfectly covering entire target object. Boundary area is missing. (d)~Background information could contain information about objects not to segment in a query image. The grass and shoes should not be segmented. \textcolor{red}{ SH: Saying "label is incorrect" and "label error" are harsh so better to say indirectly "limited information" or "beneficial to use background". }}
   \label{fig:4cases}
\end{figure}

\begin{figure}
  \centering
  
    \includegraphics[height=8.3cm]{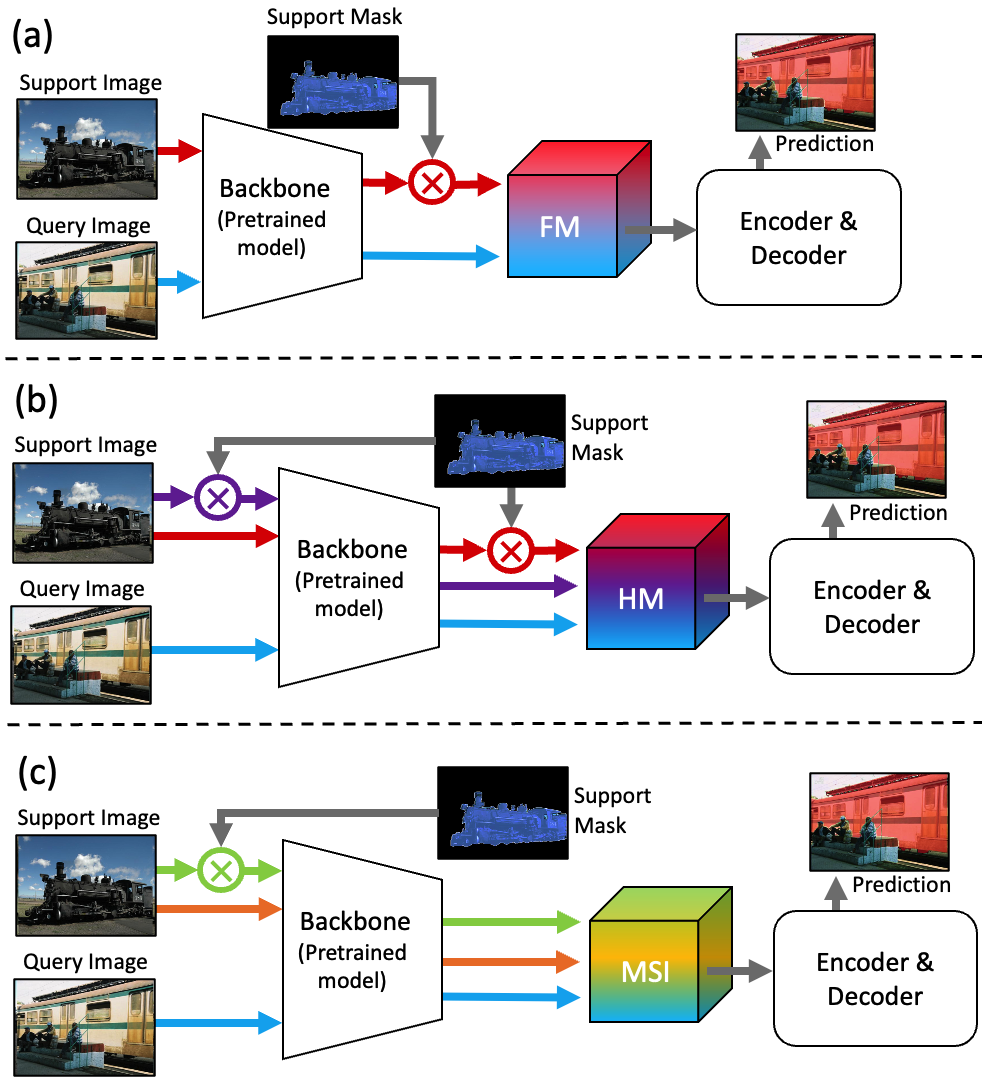}
    \caption{(a) Many FSS methods use support mask to remove background features~\cite{PMM,PANet,PFENet,FWB,CANet,ASGNet,zhang2021fewshot,HSNet, VAT2,ASNet,BAM}, denoted as feature masking (FM), and so rely only on target object features. (b) Recent work, HM~\cite{hmmasking}, merges both image masking and feature masking to achieve hybrid masking for improving target information. (c) We propose to maximize the support set information (MSI) to compensate for the limited support mask information and exploit relevant contextual information from the background.
    }

    % \comvp{Our ECCV'22 paper uses a hybrid architecture different from (a); do we want to downplay the ECCV paper or is there a point in showing (a) using either/both feature and image masking, so that it includes the proposed ECCV approach?}

    %\vspace{-0.3in}    
   % Removing background features makes a network only rely on target object features from the support set without utilizing background information. (b)~The proposed method obtains support image features~(SIF) and support target features~(STF) through two branches. STF is obtained by passing the backbone after removing the background in the support image, and SIF is obtained by passing the entire support image. }
    \label{fig:teaser}
\end{figure}

%  whole fully utilizes the available information in the entire support image.

Deep convolutional neural networks (DCNNs) have achieved state-of-the-art results across several mainstream computer vision (CV) problems, including object detection \cite{redmon2016you,ren2015faster} and semantic segmentation \cite{FCN, Deeplab, pyramidscene}. An important factor underlying the success of DCNNs is large-scale annotated datasets, which are costly and cumbersome to acquire in many dense prediction tasks, such as semantic segmentation. Moreover, these models struggle to segment novel objects when only a few annotated examples are available. Many existing few-shot segmentation (FSS) approaches \cite{Co-FCN,AMP-2,PMM,FWB,PFENet,PANet,CANet,ASGNet,DAN,FSOT,Zhang2020SGOneSG,CWT,HSNet,ASNet,BAM,VAT2,hmmasking} aim to address this shortcoming. The problem settings in FSS require accurate segmentation of a target object in a query image, given few annotated images, termed the support set, from the target class.

Shaban et al.'s work~\cite{OSLSM} 
%is believed to have introduced the FSS problem. It proposes 
introduced the first FSS model, in which a masked support image was used to extract only the target features. Using the target features, both segmentation and conditioning branches are trained to segment an object of the target class. Later, Zhang et al.~\cite{Zhang2020SGOneSG} show that extracting target features using masked average pooling~(MAP) is more beneficial for network learning than obtaining features by masking images. Since then, many FSS models~\cite{PMM,PANet,PFENet,FWB,CANet,ASGNet,zhang2021fewshot,HSNet, VAT2,ASNet,BAM} have considered MAP as the de facto technique for obtaining target features and focus on improving the encoder/decoder network. Recently, HSNet~\cite{HSNet} proposed to utilize more effective target features extracted from multiple layers of a deep backbone network. Likewise, ASNet~\cite{ASNet} and VAT~\cite{VAT2} proposed different network architectures to harness deep features.

Despite promising results, many recent methods, including HSNet~\cite{HSNet}, ASNet~\cite{ASNet}, and VAT~\cite{VAT2}, systematically struggle with few challenging FSS cases (Fig.~\ref{fig:4cases}): (a) When the support mask does not mask all instances of the same target class; (b) The support mask is unable to faithfully capture object boundaries; (c) The support mask is too small, carrying limited object information; and (d) The background contains some important contextual information, unavailable in the support mask, for accurately segmenting the target object. We conjecture that this happens because many current SOTA methods rely on support masks to completely remove the background (Fig.~\ref{fig:teaser}), which limits the useful information in several challenging FSS cases.

In this paper, we propose a new method to overcome the limited information bottleneck from the support mask. It is based on the intuition that upon maximizing the information from the masked support set images (MSI), it is possible to compensate for the limited support mask information by utilizing the important contextual information available in the typically discarded background (Fig.~\ref{fig:teaser}).
% We actualize this by maximising support set information proposing a new method, namely MSI~(see Fig.~\ref{fig:teaser}), which is an acronym for maximising support set information.
MSI jointly exploits two complementary sources of features. The first set of features is obtained by using masked support images, which only activate the target-related features in the query image. The second set of features is generated from the full support images, which activate the features of all similar objects shared between the support and query images. The former features act as an anchor for the latter in localizing a certain target class while supplementing it with the target boundary information. We summarize our key contributions as follows:

\begin{itemize}[topsep=-3pt, noitemsep]

    \item We propose MSI, an efficient and effective plug-and-play module for FSS methods. MSI harnesses masked support images to capture the 
    % detailed 
    delineated target information and exploits the entire support image for complete target information. 
    
    % \item We propose a new approach (MSI) to fully harness the entire support image in the end-to-end learning pipeline of FSS.
   
    \item We perform extensive experiments and analysis on three challenging FSS benchmarks: PASCAL-$5^i$~\cite{pascal}, COCO-$20^i$~\cite{lin2015microsoft}, and FSS-1000~\cite{FSS1000}. Results show that MSI consistently improves mIoU in the one-shot setting over all strong baselines, including HSNet~\cite{HSNet}, ASNet~\cite{ASNet}, and VAT~\cite{VAT2}.
    \item MSI improves the training speed of recent baseline models on PASCAL-5$^i$~\cite{pascal}, with 3.3x average speed-up on VAT~\cite{VAT} and 4.5x on HSNet~\cite{HSNet}.
\end{itemize}

\section{Related Works}
\label{sec:related_works}
%\vspace{-0.125in} 

\noindent\textbf{Few-shot segmentation:} OSLSM~\cite{OSLSM} is considered to be the first work introducing FSS problem. OSLSM proposed a model consisting of a condition branch and a segmentation branch. Since OSLSM, various methods have been proposed to solve the FSS problem~\cite{Co-FCN,AMP-2,PMM,FWB,PFENet,PANet,CANet,ASGNet,DAN,FSOT,Zhang2020SGOneSG,CWT,HSNet,ASNet,BAM,VAT2,hmmasking}. % co-FCN~\cite{Co-FCN} proposed a conditional network and achieved competitive accuracy. AMP~\cite{AMP-2} proposed adaptive masked proxies which builds the weights of the final segmentation layer utilizing multi-resolution imprinting. PMM~\cite{PMM} proposed to activate target objects and deactivate other objects. Khoi and Sinisa~\cite{FWB} proposed a feature weighting and boosting method to improve the discriminativeness of features.

\noindent\textbf{MAP~(Masked Average Pooling):} Zhang et al.~\cite{Zhang2020SGOneSG} proposed the MAP to collect target information from the support set. MAP masks features instead of masking support images and uses average pooling to extract target information. They argued that (1) removing the background from support images increases the variance of the input data for a unified network and (2) masking the image will make the network biased toward the target image. For these reasons, MAP was recommended to get target features. Since Zhang et al.~\cite{Zhang2020SGOneSG}, many few shot works~\cite{PMM,PANet,PFENet,FWB,CANet,ASGNet,zhang2021fewshot,HSNet, VAT2,ASNet,BAM} are following MAP to extract target features. However, by performing average pooling, spatial information is inevitably lost \cite{gap}. In this work, we utilized the masked support image to generate target features without using MAP. Therefore, we can retrieve fine-detail texture information and preserve spatial information. 

\noindent\textbf{Multi-layer features:}~Instead of using MAP, HSNet~\cite{HSNet} proposed to use deep features extracted from multiple layers and designed an effective convolution to process the features. To process the deep features effectively and efficiently, VAT~\cite{VAT} proposed a model based on the swin transformer~\cite{swin_transformer} and ASNet~\cite{ASNet} proposed the attentive squeeze network. HM~\cite{hmmasking} proposed hybrid masking to compensate for the lost details in feature masking. Although the spatial information that MAP lost was preserved, these works extracted target features counting on support masks. Therefore, when the support masks give limited information, limited target information is extracted.
% In this paper, not only the information loss when the target features are extracted by relying on the support mask is minimized but also more meaningful target features are extracted by utilizing both the entire support image and the masked support image.
In this paper, not only is information loss minimized when the target features are extracted by relying on the support mask, but also more meaningful target features are extracted by utilizing both the entire support image and the masked support image.

%   from the support set by masking background features.  However, this method still counted on support masks and focused on adding more target information to the features. 

% From our knowledge, there are no previous works 
% that enable end-to-end learning utilizing entire support image information for FSS problem.

% there are no previous studies that enabled end-to-end learning of networks using background information on FSS problems.

\noindent\textbf{Utilizing background information:} There have been several attempts to improve FSS accuracy by using background information~\cite{PMM,BAM}. PMM~\cite{PMM} proposed a method of deactivating the background and activating the target object using a duplex manner. Negative learning was carried out with objects in the background, and on the other hand, positive learning was conducted using the foreground object. Afterward, the two learning results were used to segment the target object accurately. Similarly, BAM~\cite{BAM} also proposed to use two learners. The base learner was trained using objects in the background and a meta learner was trained with the foreground object. These two learners were integrated to predict segmentation accurately. However, they assumed that the background lacks meaningful target information. Therefore, they missed target information that might exist in the background. We obtain even more target information from the background and simultaneously utilize background information to avoid segmenting wrong objects.

\noindent\textbf{Cross attention:} Both CyCTR~\cite{zhang2021fewshot} and DCAMA~\cite{DCAMA} utilize the transformer architecture to achieve cross-attention between the support and query features. This cross-attention allows the identification of target information beyond the mask area. Despite this, both methods rely on an unmasked support image to extract target information, leading to the loss of delineated detailed target information. On the contrary, MSI proposes to directly provide the model with features from the masked support image and the full support image, thereby introducing a strong inductive bias that ultimately leads to better performance. MSI computes the cosine similarity between these two complementary sets of features and concatenates them. Therefore, MSI can retain detailed target information. It is more intuitive and simpler and can be plugged seamlessly into various FSS baselines, as validated in our experiments.

\section{Overall Method}

\label{sec:method}

% The goal of CyCTR was trying to remove features that are not relevant target features. Whereas our method perform  

\begin{figure*}
  \centering
    \includegraphics[height=5.2cm]{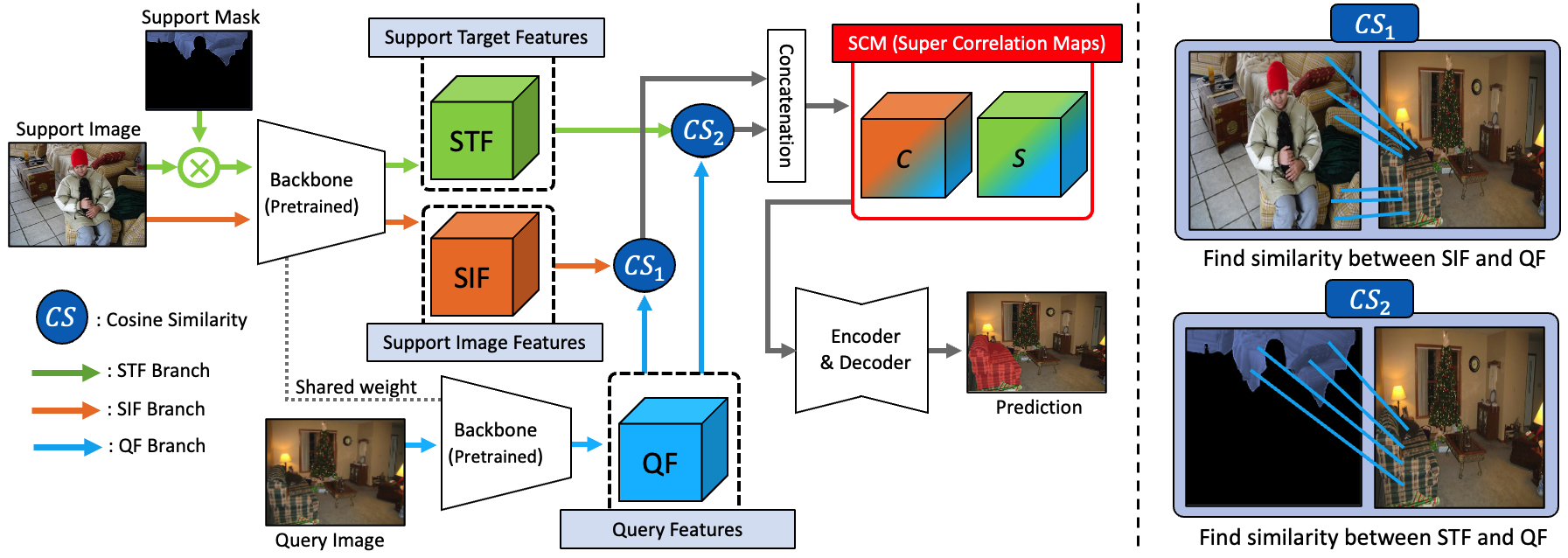}
    \caption{Overall architecture. To maximize the usability of the support set, first we extract support image features~(SIF) and support target features~(STF) through the backbone. SIF branch creates features using the entire support image, and STF branch extracts the features after removing background pixels. The query features~(QF) are obtained from the query. We obtain correlation maps by calculating the cosine similarity (CS) of the STF and SIF with the QF, respectively, and these correlation maps are concatenated to make super correlation maps~(SCM). $CS_2$ represents the cosine similarity between the target object and the query image. Therefore, only target-related information in the query image will be activated. $CS_1$, on the other hand, assesses the cosine similarity between the entire support image and the query image. Consequently, all the similar objects between the support image and the query image are activated.} 
    
    %  STF serves as guide features for SIF to help find a certain target class in the SIF and at the same time provides fine-grained target texture information that SIF does not have. STF has only information of the target class, but SIF holds the entire information of the support image. 
    % }
    \label{fig:overall}
\end{figure*}

    % \comvp{If there is a way to symmetrize the architecture layout on the lhs, I would encourage it. I.e., make the SIF and STF blocks align horizontally, place the CS1, CS2 and SCM in the "middle", then put the query elements on the right (Query Image rightmost, QF next towards the center. I understand this layout minimizes the overall area of the figure, but it is more challenging to comprehend when quickly glanced at.}

% \comvp{I would add abbreviations CS to Cosine Similarity label blocks on the right-hand side of the image, to make it clear that CS.1 and CS.2 refer to those.}
%     \comvp{I also found the way SIF and STF were connecting from the Backbone (i.e., the layout of the SIF and STF) a bit misleading; visually, it appears they come from different layers of the backbone (STF deeper, SIF shallower).  However, that is not true, from I can tell (?), as they all come from the same layers.}

Fig.~\ref{fig:overall} displays the overall architecture of our method. Following recent works~\cite{HSNet,ASNet,VAT2}, it consists of a backbone network (ResNet-50~\cite{Resnet} or ResNet-101~\cite{Resnet}) pre-trained on ImageNet~\cite{image_net} to obtain multi-layer features, and an encoder/decoder network to predict the segmentation mask.
We propose a new plug-and-play module to maximize the support set information (MSI), which compensates for limited support mask information by exploiting relevant contextual information from the background.
MSI extracts \textit{support target features}~(STF), containing delineated 
% detailed 
target class information from the support images, and \textit{support image features}~(SIF), accounting for the information in the entire support image.
Next, STF and SIF are leveraged to obtain their respective correlation maps by computing the cosine similarity with respect to the query features~(QF) obtained from the query image.
Then, the correlation maps corresponding to STF and SIF are utilized to get \textit{super correlation maps} (SCM).
Finally, this SCM is used as the input to the encoder and decoder which can be from recent FSS methods (e.g., \cite{HSNet}, \cite{VAT2}, and \cite{ASNet}).

% The core of the method is extracting features from two branches and using these to generate super correlation maps~(SCM). We extracted multi-layer features from the pretrained backbone~(ResNet50~\cite{Resnet} and ResNet101~\cite{Resnet}) following previous works~\cite{HSNet,ASNet,VAT2}. 

%We extract support target features~(STF) which have only target class information in the support image and we get support image features~(SIF) which contain the information of the entire support image. STF and SIF form correlation maps by calculating cosine similarity with query features~(QF) obtained using the query image, respectively. The correlation maps generated by using STF and SIF creates super correlation maps~(SCM) through concatenation. This SCM is used as input to the encoder and decoder used in HSNet~\cite{HSNet}, VAT~\cite{VAT2}, and ASNet~\cite{ASNet}.

\subsection{Preliminaries}
%\noindent\textbf{Preliminaries:} 
%The goal of few shot segmentation is to segment the previously unseen target class during training in a query image, $I^Q$, with a few numbers of data~(Support Set $S=\{I^S, M^S\}$) where $I^S$ is the support image and $M^S$ is the support mask). Following prior work in FSS ~\cite{HSNet,ASNet,VAT2}, we follow the episodic training learning method for training a model. An episode, $E=\{S=\{I^S, M^S\},Q=\{I^Q, M^Q\}\}$, consists of one set of support sets and one set of queries. Query mask, $M^Q$, needs to be predicted through training. Episodes become training data $D_{Tr}$ and test data $D_{Te}$, and the classes of training data and test data do not overlap. During training, episodes are repeatedly sampled from $D_{Tr}$, and for inference, episodes are sampled from $D_{Te}$.

The goal of few-shot segmentation is to train a model that can segment the target object in a query image when provided with a few annotated example images from the target class.
Following prior work in FSS ~\cite{HSNet,ASNet,VAT2}, we utilize the episodic scheme for training our model.

Specifically, we assume the availability of two disjoint sets, $C_{train}$ and $C_{test}$ as training and testing classes, respectively.
The training data $D_{train}$ is sampled from $C_{train}$ and the testing data $D_{test}$ is from $C_{test}$.
We then construct multiple episodes from $D_{train}$ and $D_{test}$. 
An episode is comprised of a support set, $S = (I^s, M^s)$, and a query set, $Q = (I^q, M^q)$, where $I^*$ and $M^*$ denote an image and its corresponding mask. 
Furthermore, $D_{train} = \{(S_i, Q_i)\}^{N_{train}}_{i=1} $ and $D_{test} = \{(S_i, Q_i)\}^{N_{test}}_{i=1}$, where $N_{train}$ represents the number of episodes for training and $N_{test}$ is the number of episodes for testing.
During training, we iteratively sample episodes from $D_{train}$ to train a model that learns a mapping from $(I^s, M^s, I^q)$ to query mask $M^q$. 
The learned model is used without further optimization by randomly sampling episodes from the testing data $D_{test}$ in the same manner and comparing the predicted query masks to the ground truth.

% \subsection{STF~(Support Target Features), SIF~(Support Image Features), and QF~(Query Features)}

\subsection{Maximizing Support-Set Information: MSI}

An important question for the FSS pipeline is where in a network it is most suitable to use masking to provide target information to the network. Earlier attempts, such as the work of Shaban et al.~\cite{OSLSM} suggested masking the background in the input image so that only target features are obtained. The rationale was that the network would supposedly be biased toward large objects, so removing the background would decrease the variance of the output parameters.
Later, Zhang et al.~\cite{Zhang2020SGOneSG} proposed to mask the background after extracting features because masking the input image makes a network biased towards the target image and changes the statistical distribution of the support image set. Many recent state-of-the-art methods have conformed to this approach~\cite{PMM,PANet,PFENet,FWB,CANet,FSOT,ASGNet,zhang2021fewshot,HSNet,VAT2,ASNet,BAM}.

\noindent\textbf{Motivation:} Although these methods report promising performance, we note that many recent methods, including HSNet~\cite{HSNet}, ASNet~\cite{ASNet}, and VAT~\cite{VAT2}, struggle to accurately predict the segmentation mask in several challenging scenarios (Fig.~\ref{fig:4cases}). We believe that this occurs because many recent FSS methods fully rely on the support mask to remove the background features (Fig.~\ref{fig:teaser}). These support masks likely present an information bottleneck in such challenging cases. For instance, when they are very small, they fail to properly encapsulate target object boundaries or do not capture all instances of the target object.

We put forth a new approach (MSI) to alleviate the aforementioned limitations of support masks. We believe that by maximizing the information from support images, it becomes possible to compensate for the limited information in the support mask and also leverage important contextual information available in the background. 
Specifically, we extract support target features~(STF), which contain delineated target class information in the support image, and support image features~(SIF), which contain contextualized information from the entire support image. 
This is because the STF enables the acquisition of more focused and fine-grained 
target information, and compared to feature masking (Fig.~\ref{fig:teaser}) STF captures more accurate object boundaries because the receptive field can cover even the area beyond target object at deep features~\cite{receptive_analysis,hmmasking}.
% which potentially gets removed when relying only on support mask~\cite{hmmasking}.  
On the other hand, SIF utilizes the full support image to capture relevant and useful target information that exists in the background. 
This allows us to obtain both detailed and complete information about the target class to avoid segmenting wrong objects.
% from the entire support image while also reducing bias toward the target class.  \honglu{I don't see how "while also reducing bias toward the target class."}
% Additionally, the STF enables the acquisition of more fine-grained target information which potentially gets removed when relying only on support mask~\cite{hmmasking}.  % Target information may exist even in the background features because the receptive field can cover even the target object area at deep features~\cite{receptive_analysis,hmmasking}. % 
% \sss{do we mean SIF here? STF doesn't have background info}\sss{isn't this what happens with input masking? we're doing feature masking, so this should not really happen if I understand correctly}
We leverage STF and SIF to obtain their respective correlation maps by computing their cosine similarity with the query features~(QF), which are obtained from the query image.
The correlation maps corresponding to STF and SIF are then combined into Super Correlation Maps (SCM), which are inputted to the encoder and decoder.
%
% Finally, this SCM is used as input to the encoder and decoder which can be from recent FSS methods \cite{HSNet}, \cite{VAT2}, \cite{ASNet}.

%To realize thise this by proposing a new method, namely MSI~(see Fig.~\ref{fig:teaser}), which is an acronym for maximising support set information.

%The proposed method utilizes both STF and SIF. Therefore, we are able to obtain information on the only target class from the support image while minimizing bias to the target class. Furthermore, by virtue of STF, we are able to get more fine detail target information which was removed by masking background features. In addition, by using SIF, it is possible to bring more information that might be missed by label error or useful for a network.
%
\noindent\textbf{Formalizing STF and SIF:} Support image features~(SIF),~$\alpha^N_{i=1}$, where $N$ is the number of features, refer to the features of the entire support image, $I^S\in\mathbb{R}^{3 \times w \times h }$, where $w$ is the width and $h$ is the height of the image. Support target features~(STF),~$\beta^N_{i=1}$, are the features from the target image, $I^T\in\mathbb{R}^{3 \times w \times h }$, which is a masked support image using support mask $\mathbf{M}^S \in \{0,1\}^{w \times h}$. Formally, %(Eq.~\ref{eq:2}).
\begin{align}  \label{eq:2}
  I^{T} & =  I^S \odot \zeta (M^S), 
\end{align}
where $\zeta ( \cdot ) $ resizes the mask $M^S$ to fit the dimension of the image $I^S$ and $\odot$ denotes the Hadamard product. Query features~(QF),~$\kappa^N_{i=1}$, are obtained from the query image, $I^Q\in\mathbb{R}^{3 \times w \times h }$.

For each image ($I^S$, $I^T$, and $I^Q$), the backbone network, $\lhd$, produces $N$ number of features ($\alpha^N_{i=1}$, $\beta^N_{i=1}$, and $\kappa^N_{i=1}$ respectively) from its intermediate layers.
\begin{align} \label{eq:3}
   {\alpha}^N_{i=1} =  \lhd(I^S),~     {\beta}^N_{i=1} =  \lhd(I^T),~
{\kappa}^N_{i=1} =  \lhd(I^Q),
\end{align}
where features, $\alpha_i,\beta_i,\kappa_i  \in\mathbb{R}^{c_i \times w_i \times h_i }$, have different sizes of channel and spatial dimensions. Features extracted from deeper layers have a large number of channels with smaller width and height dimensions.

\noindent\textbf{Super Correlation Maps (SCM):} We calculate the cosine similarities of ${\alpha}^N_{i=1}$ and ${\beta}^N_{i=1}$ with respect to ${\kappa}^N_{i=1}$ using the Eq.~\ref{eq:4} to get correlation maps $C_i \in \mathbb{R}^{ 1 \times w_i \times  h_i \times w_i \times h_i }$ and $S_i \in \mathbb{R}^{ 1 \times w_i \times  h_i \times w_i \times h_i }$. To perform multiplication between features, we reshape and transpose the features, ${\alpha}^N_{i=1}$, ${\beta}^N_{i=1}$, and ${\kappa}^N_{i=1}$ into ${\alpha'}^N_{i=1} \in\mathbb{R}^{ (w_i \cdot h_i) \times c_i }$, ${\beta'}^N_{i=1} \in\mathbb{R}^{ (w_i \cdot h_i) \times c_i }$, and ${\kappa'}^N_{i=1} \in\mathbb{R}^{ c_i \times (w_i \cdot h_i) }$.
\begin{equation}
  \begin{aligned}  \label{eq:4}
    CS_{1}={C}^N_{i=1} & = \mathrm{ReLU} \bigg(\frac{{\alpha'}^N_{i=1}\cdot {\kappa'}^N_{i=1}}{ \Vert {\alpha'}^N_{i=1} \Vert \Vert {\kappa'}^N_{i=1} \Vert  } \bigg),\\
    CS_{2}={S}^N_{i=1}  & = \mathrm{ReLU} \bigg(\frac{{\beta'}^N_{i=1}\cdot {\kappa'}^N_{i=1}}{ \Vert {\beta'}^N_{i=1} \Vert \Vert {\kappa'}^N_{i=1} \Vert  } \bigg),
  \end{aligned}
\end{equation}
where $N$ the is number of features and ReLU~\cite{relu} removes inactivated and noisy values in the correlation map $C_i$ and $S_i$. 
Correlation maps ${C}^N_{i=1}$ and ${S}^N_{i=1}$ are concatenated along the first dimension to obtain the Super Correlation Maps (SCM), ${P}^N_{i=1}  \in \mathbb{R}^{2 \times w_i \times  h_i \times w_i \times h_i }$ ~(Eq.~\ref{eq:5}).
\begin{align} \label{eq:5}
 SCM = {P}^N_{i=1} = [{C}^N_{i=1} \oplus {S}^N_{i=1}],
\end{align}
where $\oplus$ denotes the concatenation.

\subsection{Encoder and Decoder Architecture}
SCM can be fed as input to any encoder and decoder architecture of FSS methods. In order to validate our proposed method, we experiment with feeding SCM as input to three recent and strong encoder-decoder based FSS methods:
% We adopted the encoder and decoder from recent FSS models: 
% \honglu{justification on why these three methods is preferred here.}
HSNet~\cite{HSNet}, ASNet~\cite{ASNet}, and VAT~\cite{VAT}. For HSNet and VAT, the input channel size of the encoder in the models was doubled. For ASNet, SCM is merged using depth-wise attention based on Attention U-Net~\cite{attention_unet} instead of changing the input channel size. Otherwise, the architectures of the models were unchanged.

\section{Experiments}
\label{sec:result}

%\subsection{Datasets:}
\noindent\textbf{Datasets:}
We use three widely used and publicly available datasets for evaluating our method (MSI): PASCAL-$5^i$~\cite{pascal}, COCO-$20^i$~\cite{lin2015microsoft} and FSS-1000~\cite{FSS1000}.
PASCAL-$5^i$ consists of 20 classes, whereas COCO-$20^i$ has 100 classes and FSS-1000 contains 1000 classes. Following previous works~\cite{HSNet,ASNet,VAT2}, we cross-validate using 4 folds for PASCAL-$5^i$ and COCO-$20^i$ and we divide the classes into 4 groups for training and testing.
Therefore, 5 and 20 classes are used for each fold testing on PASCAL-$5^i$ and COCO-$20^i$, respectively. Other remaining classes are used for training. For FSS-1000, 1000 classes are divided into 520, 240 and 240 for the training, validation and testing. Lastly, in order to evaluate the generalizability of our method, COCO-$20^i$ is used for training, and PASCAL-$5^i$ for testing. To avoid class overlapping between training and testing, following previous works~\cite{HSNet,PFENet}, we change the PASCAL-$5^i$ class order for each fold.

%\subsection{Implementation details:}
\noindent\textbf{Implementation details:}
We incorporate MSI into three baseline models, HSNet~\cite{HSNet}, VAT~\cite{VAT}, and ASNet~\cite{ASNet}, and refer to them as HSNet~+~MSI, ASNet~+~MS and VAT~+~MSI. The backbone networks, ResNet50~\cite{Resnet} and ResNet101~\cite{Resnet}, are pre-trained on ImageNet~\cite{image_net}, and are used to extract deep features following HSNet, ASNet, and VAT~(features from conv3\_x to conv5\_x before the ReLU~\cite{relu} activation of each layer stacked to form the deep features). No fine-tuning of backbones was performed.
For a fair comparison with the existing models, we keep their default hyperparameters, as listed in their codebases. For VAT training, CATs data augmentation~\cite{data_gug_2} was used and the batch size was reduced due to GPU memory limitations. Specifically, the batch sizes 4, 8, and 4 were used for PASCAL-5$^i$~\cite{pascal}, COCO-20$^i$~\cite{lin2015microsoft}, and FSS-1000~\cite{FSS1000} respectively. No data augmentation was employed for training HSNet~+~MSI and ASNet~+~MSI.
%For training HSNet and ASNet, we follow the.

\noindent \textbf{Super-correlation maps for K-Shots$>$1:} Given a query image and K support-set images, we compute SCM for each query-support pair, resulting in K SCMs and K corresponding mask predictions from the model. All predictions are summed and normalized by the highest score~\cite{HSNet,ASNet,VAT2}.

%\subsection{Evaluation metrics.}
\noindent\textbf{Evaluation metrics:}
We report FSS performance using mean Intersection-over-Union~(mIoU) and Foreground and Background IoU~(FB-IoU), which are widely used by existing methods~\cite{FSOT,PFENet,ASGNet,HSNet,ASNet,VAT2,hmmasking}. We calculate mIoU $=  \frac{1}{n}\sum_{1}^{n}IoU$ where $n$ is the number of test cases and FB-IoU$= \frac{1}{2}{(IoU_F + IoU_B)}$ where $F$ is foreground and $B$ is background without considering classes.

% \begin{align} \label{iou}
%     IoU = \frac{Intersection}{Union} = \frac{TP}{TP+FP+FN}
% \end{align}
% where FP stands for false positives, FN for false negatives, and TP for true positives. Note that mIoU is a better metric than FB-IoU for finding the target class because mIoU measures only the target class accuracy.

\begin{table*}

\centering
\scalebox{0.8}{
\resizebox{\textwidth}{!}{%
\begin{tabular}{@{}cc|cccccc|cccccc@{}}
\toprule
\multirow{2}{*}{Backbone} & \multirow{2}{*}{Methods} & \multicolumn{6}{c|}{1-shot} & \multicolumn{6}{c}{5-shot}  
\\
 &  & $5^0$ & $5^1$ & $5^2$ & $5^3$ & mIoU & FB-IoU & $5^0$ & $5^1$ & $5^2$ & $5^3$ & mIoU & FB-IoU  \\ \midrule
\multirow{12}{*}{ResNet50~\cite{Resnet}} 

 & CWT~\cite{CWT} & 56.3 & 62.0 & 59.9 & 47.2 & 56.4 & - & 61.3 & {68.5} & 68.5 & 56.6 & 63.7 & - \\
 & RePRI~\cite{RePRI} & 59.8 & 68.3 & {62.1} & 48.5 & 59.7 & - & 64.6 & 71.4 & \textbf{71.1} & 59.3 & 66.6 & - \\
 & CyCTR~\cite{zhang2021fewshot} & {67.8} & {72.8} & 58.0 & 58.0 & 64.2 & - & {71.1} & {73.2} & 60.5 & 57.5 & 65.6 & - \\
  & BAM~\cite{BAM} & {69.0} & \textbf{73.6} & \textbf{67.6} & 61.1 & 67.8 & - & {70.6} & \textbf{75.1} & {70.8} & 67.2 & 70.9 & - \\

& DCAMA~\cite{DCAMA} & {67.5} & {72.3} & {59.6} & 59.0 & 64.6 & 76.7 & {70.3} & {73.2} & {67.4} & 67.1 & 69.5 & 80.6 \\

% 68.97 73.59 67.55 61.13 67.81 70.59 75.05 70.79 67.20 70.91 
% 67.5 72.3 59.6 59.0 64.6 (5.6) 75.7 70.5 73.9 63.7 65.8 68.5 (4.0) 79.5

 & HSNet~\cite{HSNet} & 64.3 & 70.7 & 60.3 & 60.5 & 64.0 & {76.7} & 70.3 & {73.2} & 67.4 & {67.1} & {69.5} & {80.6} \\

 & ASNet~\cite{ASNet} & {68.9} & {71.7} & {61.1} & {62.7} & {66.1}  & 77.7 & 72.6 & 74.3 & 65.3 & 67.1 & 70.8 & 80.4 \\

 & VAT~\cite{VAT} & 67.6 & 71.2 & 62.3 & 60.1 & {65.3} & {77.4} & {72.4} & {73.6} & {68.6} & {65.7} & {70.0} & {80.9} \\
% & VAT~\cite{VAT2} & 67.6 & 72.0 & 62.3 & 60.1 & {65.5} & {77.8} & {72.4} & {73.6} & {68.6} & {65.7} & {70.1} & {80.9} \\

  \cmidrule(l){2-14} 

& HSNet~+~MSI & 68.1 & {71.5} & 58.2 & {62.9} & {65.2} & 76.5  & 70.7 & 72.8 & 61.5 & 66.6 & 67.9  & 78.2 \\

& ASNet~+~MSI & {69.2} & {71.7} & {59.7} & {64.4} & {66.3}  & 77.9 & 72.0 & 73.2 & 64.0 & 68.0 & 69.3 & 80.2 \\

& VAT~+~MSI & \textbf{71.0} & {72.5} & {63.8} & \textbf{65.9} & \textbf{68.3} & \textbf{79.1}  & \textbf{73.0} & {74.2} & 66.6 & \textbf{70.5} & \textbf{71.1}   & \textbf{81.2}  \\

  \midrule

 & CWT~\cite{CWT} & 56.9 & 65.2 & 61.2 & 48.8 & 58.0 & - & 62.6 & 70.2 & \textbf{68.8} & 57.2 & 64.7 & - \\
 & RePRI~\cite{RePRI} & 59.6 & 68.6 & {62.2} & 47.2 & 59.4 & - & 66.2 & 71.4 & 67.0 & 57.7 & 65.6 & -  \\
 & CyCTR~\cite{zhang2021fewshot} & {69.3} & {72.7} & 56.5 & 58.6 & 64.3 & 72.9 & {73.5} & 74.0 & 58.6 & 60.2 & 66.6 & 75.0 \\

& DCAMA~\cite{DCAMA} & {65.4} & {71.4} & {63.2} & 58.3 & 64.6 & 77.6 & {70.7} & {73.7} & {66.8} & 61.9 & 68.3 & 80.8 \\  

% 65.4 71.4 63.2 58.3 64.6 (4.7) 77.6 70.7 73.7 66.8 61.9 68.3 (4.4) 80.8
  
\multirow{4}{*}{ResNet101~\cite{Resnet}} 
 & HSNet~\cite{HSNet} & 67.3 & {72.3} & 62.0 & 63.1 & {66.2} & 77.6 & 71.8 & {74.4} & 67.0 & {68.3} & {70.4} & {80.6} \\

 & ASNet~\cite{ASNet} & {69.0} & {73.1} & {62.0} & {63.6} & {66.9}  & 78.0 & 73.1 & 75.6 & 65.7 & 69.9 & 71.1 & 81.0 \\

& VAT~\cite{VAT} & {68.4} & 72.5 & \textbf{64.8} & {64.2} & {67.5}  & {78.8} & {73.3} & {75.2} & {68.4} & {69.5} & {71.6} & {82.0} \\
% & VAT~\cite{VAT2} & {70.0} & 72.5 & \textbf{64.8} & {64.2} & {67.9}  & {79.6} & \textbf{75.0} & {75.2} & {68.4} & {69.5} & {72.0} & \textbf{83.2} \\

  \cmidrule(l){2-14}

  & HSNet~+~MSI  & {70.5} & 72.9 & 60.6 & {64.3} & {67.1} & {77.8} & {71.9} & 74.9 & 64.1 & 67.7 & 69.7 & 79.5 \\

  & ASNet~+~MSI & {70.5} & {73.8} & {61.3} & {65.5} & {67.8}  & {78.8} & {73.4} & {75.5} & 66.2 & 71.0 & 71.5 & 81.3 \\

& VAT~+~MSI  & \textbf{73.1} & \textbf{73.9} & {64.7} & \textbf{68.8} & \textbf{70.1}  & \textbf{82.3} & \textbf{73.6} & \textbf{76.1} & 68.0 & \textbf{71.3} & {\textbf{72.2}}  & \textbf{82.3} \\

% 73.14, 73.27. 64.70. 68.81. 69.98

 \bottomrule

\end{tabular}
}
}
\caption{Performance evaluation on Pascal-5$^i$~\cite{pascal}. Best results are shown in \textbf{bold}.} %An explanation of the performance degradation can be found in Section 5.}
\label{table:performance_pascal}

\end{table*}

\begin{table*}
  \centering
    \begin{minipage}{.60\textwidth}
\centering
\resizebox{\textwidth}{!}{%
\begin{tabular}{@{}cc|ccccc|ccccc@{}}
\toprule
\multirow{2}{*}{Backbone} & \multirow{2}{*}{Methods} & \multicolumn{5}{c|}{1-shot} & \multicolumn{5}{c}{5-shot} 
\\
 &  & $20^0$ & $20^1$ & $20^2$ & $20^3$ & mIoU & $20^0$ & $20^1$ & $20^2$ & $20^3$ & mIoU \\ \midrule
\multirow{10}{*}{ResNet50~\cite{Resnet}} 

 & RePRI~\cite{RePRI} & 32.0 & 38.7 & 32.7 & 33.1 & 34.1  & 39.3 & 45.4 & 39.7 & 41.8 & 41.6  \\
& CyCTR~\cite{zhang2021fewshot} & {38.9} & 43.0 & {39.6} & {39.8} & {40.3}  & 41.1 & 48.9 & 45.2 & {47.0} & 45.6  \\

& BAM~\cite{BAM} & {43.4} & \textbf{50.6} & {47.5} & {43.4} & {46.2}  & \textbf{49.3} & 54.2 & 51.6 & {49.6} & 51.2  \\

& DCAMA~\cite{DCAMA} & {41.9} & {45.1} & {44.4} & {41.7} & {43.3}  & 45.9 & 50.5 & 50.7 & {46.0} & 48.3  \\

% 41.9 45.1 44.4 41.7 43.3 (1.5) 69.5 45.9 50.5 50.7 46.0 48.3 (2.3) 71.7

 & HSNet~\cite{HSNet} & 36.3 & {43.1} & 38.7 & 38.7 & 39.2  & {43.3} & {51.3} & {48.2} & {45.0} & {46.9}  \\

& ASNet~\cite{ASNet} & {41.5} & 44.1 & 42.8 & {40.6} & {42.2}   & {47.6} & {50.1} & 47.7 & {46.4} & {47.9}  \\

& VAT~\cite{VAT} & {39.0} & 43.8 & 42.6 & {39.7} & {41.3} & {44.1} & 51.1 & 50.2 & 46.1 & 47.9  \\

 \cmidrule(l){2-12}

& HSNet~+~MSI & {40.9} & {46.9} & {48.8} & {45.3} & {45.5}   & {45.3}  & {53.5} & {53.1}  & {49.3}  & {50.3}  \\ 

 & ASNet~+~MSI & {41.5} & {46.3} & {43.5} & {42.1} & {43.4}  & 46.0 & 50.7 & 47.5 & {46.9} & {47.8} \\

& VAT~+~MSI & \textbf{42.4} & {49.2} & \textbf{49.4} & \textbf{46.1} & \textbf{46.8}  & 47.1 & \textbf{54.9} & \textbf{54.1} & \textbf{51.9} & \textbf{52.0}  \\

 \midrule

\multirow{6}{*}{ResNet101~\cite{Resnet}} 

& DCAMA~\cite{DCAMA} & {41.5} & {46.2} & {45.2} & {41.3} & {43.5}  & {48.0} & 58.0 & 54.3 & {47.1} & 51.9  \\

% 41.5 46.2 45.2 41.3 43.5 (2.2) 69.9 48.0 58.0 54.3 47.1 51.9 (4.5) 73.3

& HSNet~\cite{HSNet} & 37.2 & {44.1} & {42.4} & {41.3} & {41.2} & {45.9} & {53.0} & {51.8} & {47.1} & {49.5} \\

& ASNet~\cite{ASNet} & {41.8} & 45.4 & 43.2 & {41.9} & {43.1} & {48.0} & 52.1 & 49.7 & 48.2 & {49.5} \\

& VAT~\cite{VAT} & 39.5 & 44.4 & 46.1 & 40.4 & 42.6 & 45.2 & 54.1 & 51.1 & 47.1 & 49.4 \\

\cmidrule(l){2-12}

& HSNet~+~MSI & {42.4} & {50.1} & {49.5} & \textbf{48.3} & {47.6} & {48.0}  & {57.3} & {52.6}  & {52.6}  & {52.6} \\ 

 & ASNet~+~MSI & {42.9} & 45.2 & {44.3} & {43.4} & {44.0} & {47.7} & {49.9} & 49.0 & {48.8} & {48.9} \\

& VAT~+~MSI & \textbf{44.8} & \textbf{54.2} & \textbf{52.3} & 48.0 & \textbf{49.8} & \textbf{49.3} & \textbf{58.0} & \textbf{56.1} & \textbf{52.7} & \textbf{54.0}\\

  \bottomrule
  
\end{tabular}
}
\caption{Performance evaluation on COCO-20$^i$~\cite{lin2015microsoft}. Best results are shown in \textbf{bold}.}
\label{table:performance_coco}
    \end{minipage}
    \begin{minipage}{.27\textwidth}
\centering

\resizebox{\textwidth}{!}{

\begin{tabular}{@{}cc|cc@{}}
\toprule
\multirow{2}{*}{Backbone} & \multirow{2}{*}{Methods} & \multicolumn{2}{c}{mIoU} \\
 &  & 1-shot & 5-shot \\ \midrule
\multirow{5}{*}{ResNet50~\cite{Resnet}} 
& HSNet~\cite{HSNet} & {85.5} & 87.8 \\

  & VAT~\cite{VAT} & {{89.5}} & {90.3} \\
 % & VAT~\cite{VAT2} & \textbf{ {90.1}} & \textbf{90.7} \\
  
\cmidrule(lr){2-4}

 & HSNet~+~MSI & 87.5 & 88.4 \\ 

& VAT~+~MSI & \textbf{90.0} & \textbf{90.6} \\ 

\midrule

\multirow{5}{*}{ResNet101~\cite{Resnet}} & HSNet~\cite{HSNet} & 86.5 & 88.5 \\

& VAT~\cite{VAT} & {90.0}  & {90.6} \\

\cmidrule(lr){2-4}

% & VAT~\cite{VAT2} & {90.3}  & {90.8} \\
  
& HSNet~+~MSI & 88.1 & 89.2 \\

& VAT~+~MSI & \textbf{90.6} & \textbf{91.0}\\ 

 \bottomrule
\end{tabular}
}
\caption{Performance evaluation on FSS-1000~\cite{FSS1000}.}
\label{table:performance_fss}
    \end{minipage}

%\vspace{-0.1in}
\end{table*}

\begin{table}[!htp]
\centering
\scalebox{0.6}{%
\begin{tabular}{@{}cc|c|c@{}}
\toprule
\multirow{2}{*}{Backbone} & \multirow{2}{*}{Methods} & \multicolumn{1}{c|}{1-shot} & \multicolumn{1}{c}{5-shot}  \\
&  &  mIoU &  mIoU    \\ \midrule
\multirow{4}{*}{ResNet50~\cite{Resnet}} 
 & HSNet~\cite{HSNet} & 61.6  & {68.7} \\
 
   & VAT~\cite{VAT} & 64.5 & 69.7 \\

 \cmidrule(l){2-4}

    & HSNet~+~MSI & {66.0} & {70.8} \\  

  & VAT~+~MSI&  \textbf{67.8} & \textbf{72.6} \\  
  
   \midrule
  
\multirow{4}{*}{ResNet101~\cite{Resnet}}  & HSNet~\cite{HSNet} & {64.1} & {70.3}\\

   & VAT~\cite{VAT} & 66.8 & 71.4 \\

 \cmidrule(l){2-4}

   & HSNet~+~MSI&  {67.3} &  {72.3} \\  

  & VAT~+~MSI&  \textbf{69.2} & \textbf{74.1} \\  
  
 \bottomrule

\end{tabular}
}
\caption{Generalizability performance evaluation on PASCAL-5$^i$~\cite{pascal} after training with COCO-20$^i$~\cite{lin2015microsoft}. Best results are shown in \textbf{bold}.}
\label{table:performance_pascal_shift}

%\vspace{-0.3in}
\end{table}

% MSI was instantiated into three strong baseline models HSNet~\cite{HSNet}, ASNet~\cite{ASNet}, and VAT~\cite{VAT2} and we name them as HSNet~+~MSI, ASNet~+~MSI, and VAT~+~MSI, respectively. \sss{This was already explained in ``Implementation details'' (line 456)}

\subsection{Results}
\label{subsection:Results}
\noindent\textbf{PASCAL-$5^i$~\cite{pascal}:} Tab.~\ref{table:performance_pascal} compares the results of the proposed method~(MSI) with baseline models on PASCAL-$5^i$. MSI allowed almost all experiments to set a new SOTA record. We observed that VAT~+~MSI provided the most noticeable gains. In the 1-shot test, VAT~+~MSI provided a 3.0\% gain with ResNet50 and a 2.5\% gain with ResNet101.

% In the 1-shot test, HSNet~+~MSI with ResNet50~\cite{Resnet} improved by 1.2\% in mIoU compared to the HSNet and by 0.9\% in mIoU with ResNet101~\cite{Resnet}. 
% %
% %In the 5-shot test, the mIoU performance was inferior to HSNet. ASNet~+~MSI showed a similar pattern to HSNet~+~MSI. 
% In the 1-shot test, ASNet~+~MSI with ResNet50~\cite{Resnet} improved by 0.2\% and with ResNet101~\cite{Resnet}, the mIoU improved by 0.9\%. We also observed that VAT~+~MSI provided noticeable gains in both 1-shot and 5-shot tests, and the performance improvement for 1-shot was particularly significant. In the 1-shot test, VAT~+~MSI provided a 2.8\% gain with ResNet50~\cite{Resnet} and a 2.1\% gain with ResNet101~\cite{Resnet} in mIoU. 
% %In the 5-shot test, VAT~+~MSI delivered 1.0\% and 0.2\% gains in mIoU with ResNet50~\cite{Resnet} and ResNet101~\cite{Resnet} respectively.

\noindent\textbf{COCO-$20^i$~\cite{lin2015microsoft}:}
Tab.~\ref{table:performance_coco} reports results on the COCO-$20^i$ dataset.
% HSNet~\cite{HSNet}, ASNet~\cite{ASNet}, and VAT~\cite{VAT} are selected and compared with the proposed method~(see Tab.~\ref{table:performance_coco}). 
%
% Similar to PASCAL-$5^i$, we compared three baselines, HSNet~\cite{HSNet}, ASNet~\cite{ASNet}, and VAT~\cite{VAT}, before and after integrating MSI. 
%ASNet~+~MSI showed a slight decrease in performance for 5-shot; however for 1-shot, with ResNet50~\cite{Resnet} we observed a 1.2\% mIoU improvement over ASNet, and with ResNet101~\cite{Resnet}, the improvement was 0.9\%. 
Our method delivered consistent improvement in almost all experiments. VAT~+~MSI provided a gain of 5.5\% with ResNet50~\cite{Resnet} and 7.2\% with ResNet101~\cite{Resnet} in mIoU in the 1-shot test, and delivered gains of 5.8\% and 5.6\% in the 5-shot test. 
%
% HSNet~+~MSI improved by 6.3\% with ResNet50~\cite{Resnet} and 6.5\% with ResNet101~\cite{Resnet} in the 1-shot test. Moreover in the 5-shot, HSNet~+~MSI improved by 3.4\% and 3.1\% in mIoU with ResNet50~\cite{Resnet} and Resnet101~\cite{Resnet}, respectively.

\noindent\textbf{FSS-1000~\cite{FSS1000}:}
Tab.~\ref{table:performance_fss} compares the performance of our MSI on FSS-1000 using two baselines: HSNet~\cite{HSNet} and VAT~\cite{VAT}. HSNet~+~MSI delivered 2.0\% and 0.6\% mIoU improvements on 1-shot and 5-shot tests, respectively, with ResNet50~\cite{Resnet}. HSNet~+~MSI  with ResNet101~\cite{Resnet} showed 1.6\% and 0.7\% gains in mIoU. VAT~+~MSI showed gains of 0.5\% and 0.3\% in mIou with ResNet50~\cite{Resnet} and gains of 0.3\% and 0.2\% in mIoU with ResNet101~\cite{Resnet}.

\noindent\textbf{Generalization Test:} We tested the generalizability of trained models in Tab.~\ref{table:performance_pascal_shift}. 
HSNet~\cite{HSNet} and VAT~\cite{VAT} were compared with HSNet~+~MSI and VAT~+~MSI, respectively. Both HSNet~+~MSI and VAT~+~MSI delivered significant mIoU gains on 1-shot and 5-shot tests. HSNet~+~MSI with ResNet50~\cite{Resnet} provided 4.4\% and 2.1\% improvements and VAT~+~MSI showed 3.4\% and 2.7\% gains. %Results suggest leveraging
% our method
%MSI can improve the generalizability of models.

% ~All models were trained on COCO-20$^i$~\cite{lin2015microsoft} and then tested on PASCAL-5$^i$~\cite{pascal}. 

\begin{figure*}[!htp]
  \centering
    \includegraphics[height=6.9cm]{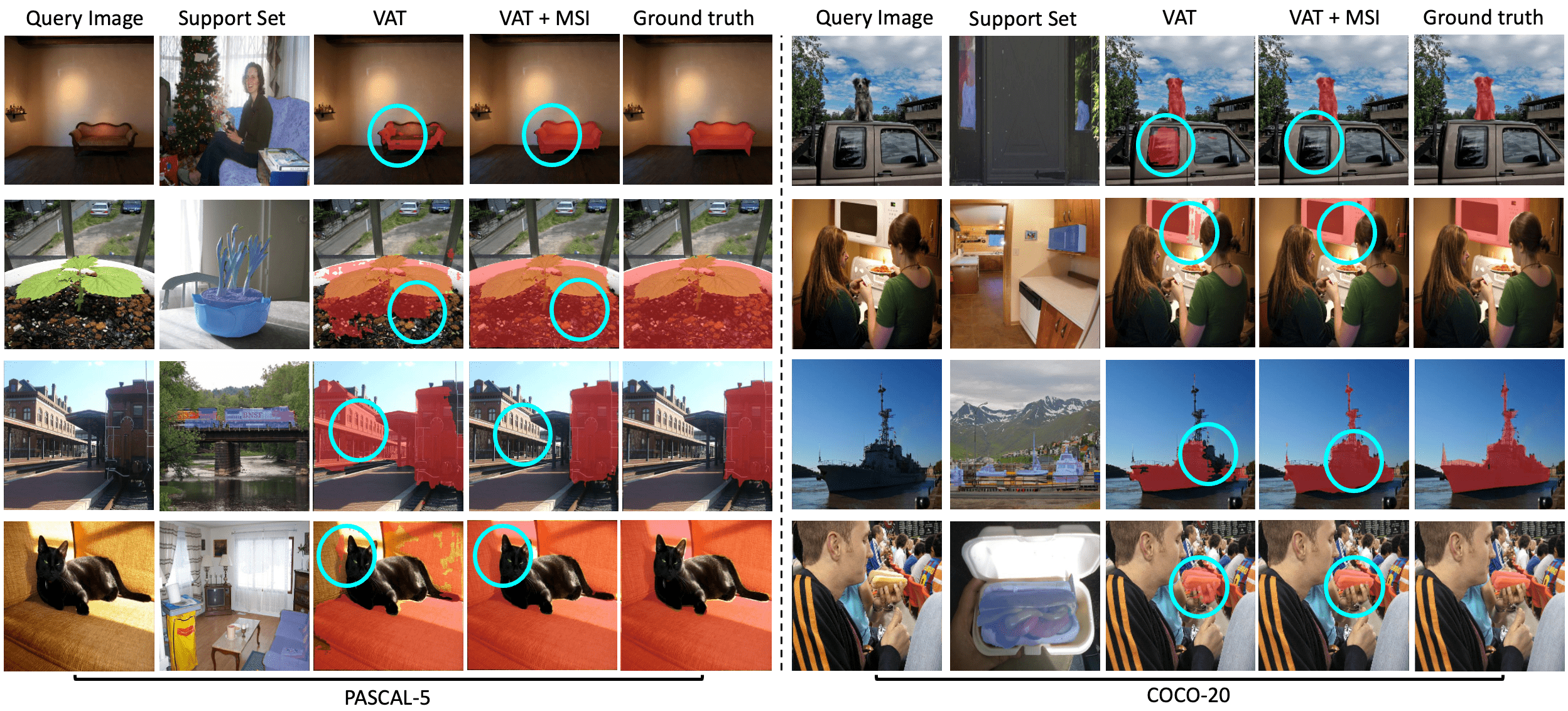}
    \caption{Visual comparison of VAT~\cite{VAT} and VAT~+~MSI with ResNet50~\cite{Resnet} on PASCAL-5$^i$~\cite{pascal} and COCO-20$^i$~\cite{lin2015microsoft}.
    }
    
    \label{fig:visual_comparison}
\end{figure*}

\noindent\textbf{Qualitative Results:} According to the visual comparison, MSI allows VAT to capture, detailed, accurate, and complete object information, and thus lead to an improved performance and avoid segmenting the wrong~object~(Fig.~\ref{fig:visual_comparison}).

\noindent\textbf{Small improvement with ASNet/HSNet:} Performance drop in 5-shot is observed on PASCAL with HSNet and ASNet. There could be two reasons. First, both HSNet and ASNet models are limited in fully harnessing MSI’s potential. Particularly, ASNet downsamples support features by pooling which loses target information and hence limits the capability of both support target features~(STF) and support image features~(SIF) in MSI. HSNet lacks self-attention, which would aid MSI by facilitating the network in finding target information from the entire SIF by leveraging STF. Second, ASNet, HSNet, VAT, and their MSI-empowered counterparts are trained for the one-shot setting, which might prevent MSI from specializing in 5-shot testing~(Tab~\ref{table:performance_pascal}).

\begin{table}
\centering
\scalebox{0.77}{%
 \begin{tabular}{c | c c c | c c} 
 \hline
 Method \& Backbone & FM & STF & SIF & mIoU & FB-IoU \\
 \hline
  & &  & \checkmark & 60.7 & 72.0 \\ 
 & \checkmark &   &  & 65.3 (Baseline~\cite{VAT}) & 77.4 \\
 &  & \checkmark &  & 65.2 &  77.4\\
VAT~(ResNet50) & \checkmark & \checkmark &  & 66.3~(HM~\cite{hmmasking}) & 78.0 \\
 &  & \checkmark & \checkmark & \textbf{68.3} (Ours) & \textbf{79.1} \\
 & \checkmark &  & \checkmark & 65.7 & 77.5 \\
 & \checkmark & \checkmark & \checkmark  & 67.1 & 78.8 \\
 \hline
 
 \end{tabular}
 }
\caption{\small Ablation study of the different combination of features with VAT~\cite{VAT}~(ResNet50~\cite{Resnet}) on PASCAL-5$^i$~\cite{pascal}). \checkmark denotes that these features are utilized to generate super correlation maps~(SCM). The best results are shown in \textbf{bold}.}
 
\label{table:Ablation_study1}
\end{table}

\begin{table}
  \centering
    \begin{minipage}{.47\textwidth}
\centering
\resizebox{\textwidth}{!}{
\begin{tabular}{@{}cc|ccccc}
\toprule
\multirow{2}{*}{Backbone} & \multirow{2}{*}{Methods} & \multicolumn{5}{c}{1-shot} 
\\
 &  & $20^0$ & $20^1$ & $20^2$ & $20^3$ & mIoU   \\ \midrule
\multirow{5}{*}{ResNet50~\cite{Resnet}} 
& VAT~\cite{VAT}~(Baseline) & 67.6 & 71.2 & 62.3 & 60.1 & {65.3} \\
& VAT+MSI~(Feature Add.)& 68.5 & 71.0 & 60.8 & 62.0 & 65.6 \\
 & VAT+MSI~(Correlation Add.) & 67.0 & 68.2 & 53.5 & 56.0 & 61.2 \\
 
  & VAT+MSI~(Attention) & 69.9 & \textbf{72.5} & 63.4 & 63.8  & 67.4 \\

 & VAT+MSI~(Ours) & \textbf{71.0} & \textbf{72.5} & \textbf{63.8} & \textbf{65.9} & \textbf{68.3} \\

\bottomrule 
\end{tabular}

}

    \end{minipage}
    \caption{\small Different methods of generating super correlation maps~(SCM) on PASCAL-5$^i$~\cite{pascal}. Best results are shown in \textbf{bold}. }
\label{table:Ablation_study2}

  \end{table}

\begin{table}
  \centering
    \begin{minipage}{.47\textwidth}
\centering
\resizebox{\textwidth}{!}{
\begin{tabular}{@{}cc|ccccc}
\toprule
\multirow{2}{*}{Backbone} & \multirow{2}{*}{Methods} & \multicolumn{5}{c}{1-shot} 
\\
 &  & $20^0$ & $20^1$ & $20^2$ & $20^3$ & mIoU   \\ \midrule
\multirow{2}{*}{ResNet50 (PASCAL-5$^i$)} 
 & VAT~\cite{VAT} & 59.9 & 42.4 & 42.0 & 45.5 & 47.5 \\
& VAT~+~MSI & \textbf{68.1} & \textbf{46.6} & \textbf{43.6} & \textbf{47.1} & \textbf{51.4}  \\
 
 \midrule 
\multirow{2}{*}{ResNet50 (COCO-20$^i$)}  
& VAT~\cite{VAT} & 29.7 & 32.4 & 36.0 & 29.0 & 31.8 \\
 & VAT~+~MSI & \textbf{30.7} & \textbf{37.9} & \textbf{40.9} & \textbf{33.9} & \textbf{35.9} \\

\bottomrule 
\end{tabular}

}

    \end{minipage}
    \caption{\small Performance comparison on PASCAL-5$^i$~\cite{pascal} and COCO-20$^i$~\cite{lin2015microsoft} for cases where support masks occupy below 5\% of the support images. Best results are shown in \textbf{bold}. }
\label{table:Ablation_study3}

  \end{table}

\begin{table}
  \centering
    \begin{minipage}{.47\textwidth}
\centering
\resizebox{\textwidth}{!}{
\begin{tabular}{@{}cc|ccccc}
\toprule
\multirow{2}{*}{Backbone} & \multirow{2}{*}{Methods} & \multicolumn{5}{c}{1-shot} 
\\
 &  & $20^0$ & $20^1$ & $20^2$ & $20^3$ & mIoU   \\ \midrule
\multirow{4}{*}{ResNet50~\cite{Resnet}} 
&VAT~(w/o person in BG) & 67.7 & 70.9 & 61.3 & 61.6 & 65.2  \\ 
& VAT~(w/ person in BG) & {67.6} & {71.2} & {62.3} & {60.1} & {65.3} \\ 

% 66.8, 70.4, 61.3, 61.0

&VAT~+~MSI~(w/o person in BG) & 69.4 & 71.4 & 60.5 & 64.1 & 66.3  \\ 
& VAT~+~MSI~(w/ person in BG) & \textbf{71.0} & \textbf{72.5} & \textbf{63.8} & \textbf{65.9} & \textbf{68.3} \\ 

\bottomrule 
\end{tabular}

}

    \end{minipage}
    \caption{\small Training VAT~+~MSI on PASCAL-5$i$\cite{pascal} without the person class existing as background of the support image.}

\label{table:Ablation_study4}

  \end{table}

% \subsection{Why MSI is useful?}
\subsection{When is MSI useful?}
% Fig.~\ref{fig:4cases} draws visual comparisons between VAT~+~MSI and the baseline VAT~\cite{VAT}, which depends only on target features. 
% Tab.~\ref{table:Ablation_study1} highlights the superiority of the VAT~+~MSI, which utilizes the entire features from the support image. These results indicate that using the entire support image improves results compared to using only the target information. Now, 
As MSI can better capture target object information, it is more capable of handling challenging FSS scenarios. 
Fig.~\ref{fig:4cases} draws visual comparisons between VAT~+~MSI and the baseline VAT~\cite{VAT} on four challenging FSS cases where MSI is particularly effective. We discuss these 
% four 
% challenging 
cases below.
% We describe four challenging FSS cases where MSI is effective (Fig.~\ref{fig:4cases}). 

%The usefulness of the proposed method is suggested by four reasons in Fig.~\ref{fig:4cases}. The following cases were investigated to deeply understand the usefulness of the method.

\begin{figure}[!htp]
  \centering
    \includegraphics[height=1.6cm]{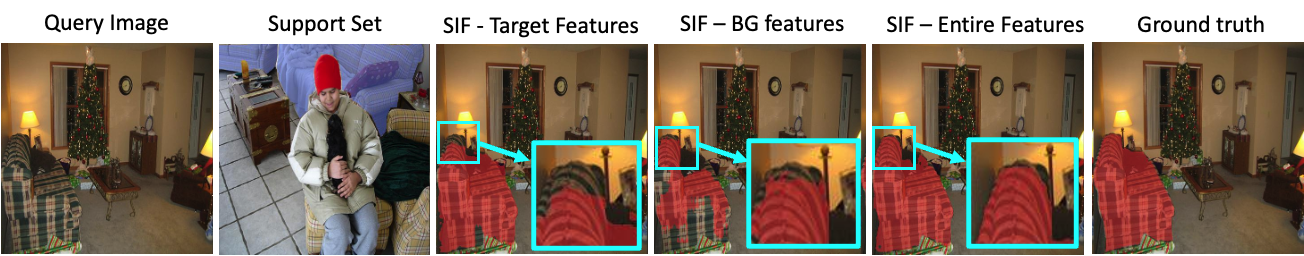}

   \caption{Comparison of VAT~+~MSI on PASCAL-5$^i$~\cite{pascal} with ResNet50~\cite{Resnet}~using different kinds of features in SIF: SIF with only target features, SIF with only background features, and SIF with the entire image's features.}% The support mask fails to mask the same class, sofa on the right side. Therefore, SIF with target features are generated only from the upper sofa. The SIF with background features are also not much effective overall (in predicting the accurate mask) even though they better segment the magnified part in cyan box. However, SIF with entire image features, which incorporate both the former and the latter features, allows accurate segmentation of the target object, magnified in cyan box and otherwise.}

%   Visual comparison of VAT~+~MSI on PASCAL-5$^i$~\cite{pascal}~(ResNet50~\cite{Resnet}) using different features in SIF. Support mask failed to mask the same class, sofa on the right side. Therefore, target features are generated only from the upper sofa. The performance of utilizing only target features is inferior to using entire features because the target features give limited information to a model. In general, the background features are not as effective as target features but the background features showed better performance in the cyan box.}
   \label{fig:features}
\end{figure}

\noindent\textbf{When an instance of the target class is not masked:} 
% Fig.~\ref{fig:features} draws visual comparison of VAT~+~MSI on PASCAL-5$^i$~\cite{pascal}~(ResNet50~\cite{Resnet}) using different kinds of features in SIF: SIF with only target features, SIF with only background features, SIF with entire image features. 
Fig.~\ref{fig:features} visualizes different kinds of features in SIF using VAT~+~MSI on PASCAL-5$^i$~\cite{pascal}~(ResNet50~\cite{Resnet}).
% : SIF with only target features, SIF with only background features, SIF with the entire image's features. 
% We observe that the support mask fails to mask the same class, sofa on the right side. Therefore, SIF with target features are generated only from the upper sofa. 
We observe that the support mask fails to mask an instance of the same class, i.e., the sofa on the right side. Therefore, SIF with target features is generated only from the upper sofa.
Next, SIF with background features is not very effective overall, even though they better segment the magnified part in the cyan box.
%
% Finally, SIF with the entire image's features, which incorporate both the former and the latter features, allows accurate segmentation of the target object, magnified in the cyan box and otherwise.
Finally, SIF with the entire image's features (ours), which incorporate both the former and the latter features, allows accurate segmentation of the target object, both inside and outside the cyan box. This indicates that background features have target information and utilizing the features improves the performance of FSS.

% features but the background features showed better performance in the cyan box.

% %The performance of utilizing only target features is inferior to using entire features because the target features give limited information to a model. 

% In general, the background features are not as effective as target features but the background features showed better performance in the cyan box.

% We observe that the missing target features are used for target prediction~(Fig.~\ref{fig:features}). The performance using only the target features is inferior to using entire image features. This indicates that VAT~+~MSI utilizes the important contextual information from the background features to segment the target class. Also, we test the model with only background features. The result shows that in general the performance is inferior to using the target features but background features do help capturing different information which is missed by the target features.

% After training VAT~+~MSI, the background features were masked in SIF to leave only the target features. 

% Fig.~\ref{fig:visual_comparison} shows the visual performance comparison between VAT~+~MSI and baseline VAT that depends only on target features. Tab.~\ref{table:Ablation_study1} shows the superiority of the VAT~+~MSI method using the background features. These results indicate that using the entire support image brought better results than using only the target information.

\noindent\textbf{Support mask is very small:} 
In PASCAL-5 and COCO-20, there are several cases where the support masks are very small because the target object only occupies a small part of the support image. 
% To show that MSI is more effective than the existing methods on very small target objects, in both PASCAL-5$^i$~\cite{pascal} and COCO-20$^i$~\cite{lin2015microsoft}, we compare the mIoU for cases when the support mask occupies less than 5\% of the support image. Compared to VAT~\cite{VAT}, VAT~+~MSI displays noticeable gains of 3.9\% and 4.1\% (in mIOU) for PASCAL-5$^i$ and COCO-20$^i$, respectively~(Tab.~\ref{table:Ablation_study3}).
% To show that MSI is more effective than the existing methods on very small target objects, in both PASCAL-5$^i$~\cite{pascal} and COCO-20$^i$~\cite{lin2015microsoft}, 
We compare the mIoU for cases when the support mask occupies less than 5\% of the support image~(Tab.~\ref{table:Ablation_study3}). Compared to VAT~\cite{VAT}, VAT~+~MSI displays noticeable gains of 3.9\% and 4.1\% (in mIOU) for PASCAL-5$^i$ and COCO-20$^i$, respectively. These results show that MSI is more effective than the existing methods on very small target objects in both PASCAL-5$^i$~\cite{pascal} and COCO-20$^i$~\cite{lin2015microsoft}.

%the mIoU of the results of the support mask occupying less than 5\% of the support image (see Tab.~\ref{table:Ablation_study3}). The VAT~+~MSI method showed great improvement in both PASCAL-5$^i$ and COCO-20$^i$ compared to VAT~\cite{VAT}. This indicates MSI can achieve good performance even with limited target information.

\begin{figure}[t]
  \centering
    \includegraphics[height=2.05cm]{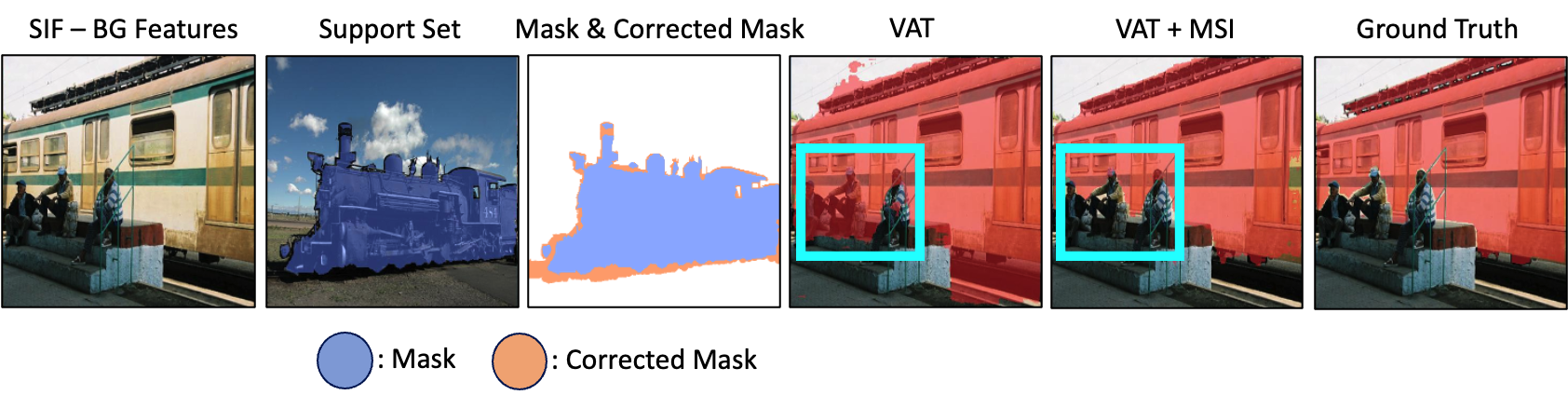}

   \caption{Missing information around boundary on PASCAL-5$^i$~\cite{pascal}. The mask is not perfect to cover the entire target object.}
   \label{fig:boundary_error}
\end{figure}

\noindent\textbf{Support mask is missing some of the target boundary:}  We observe that in PASCAL-5$^i$~\cite{pascal}, COCO-20$^i$~\cite{lin2015microsoft}, and even FSS-1000~\cite{FSS1000}, the masks around boundaries can be imperfect (Fig.~\ref{fig:boundary_error}). By masking the features with an inaccurate mask, the boundary information of the target disappears inadvertently (Fig.~\ref{fig:boundary_error}). Therefore, compared to ours, previous methods face difficulties in finding the target in the query image with only limited target information.

\begin{figure}[t]
  \centering
    \includegraphics[height=2cm]{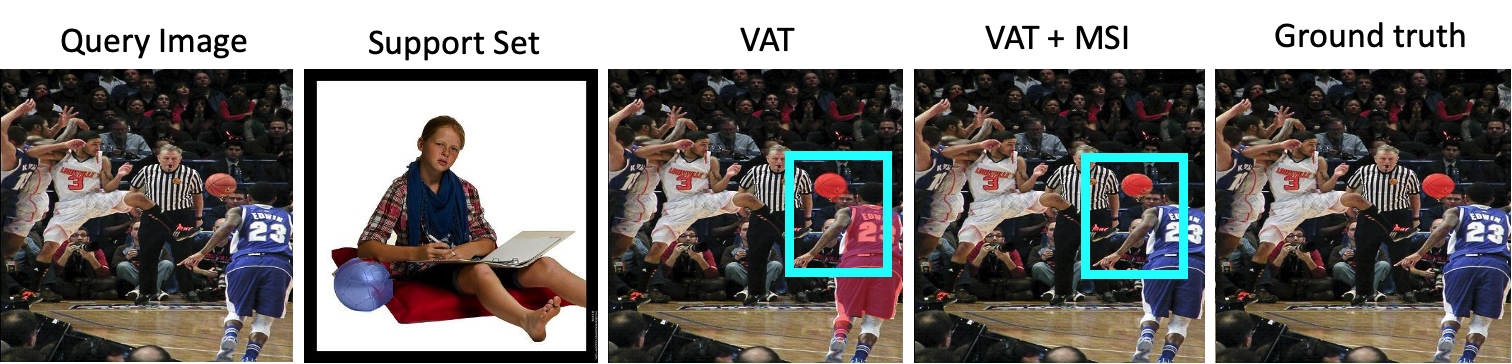}

   \caption{Comparison of performance when background information of support set is relevant to the background in the query image. Note that the person is in the background of the support image (target object is the ball). VAT failed to segment the ball correctly, while VAT~+~MSI was able to differentiate the ball and people.}
   % accurately.}
   %on COCO-20~$^i$~\cite{lin2015microsoft} when background information of support set and query image is similar to each other.}
   \label{fig:background_info}
\end{figure}

\noindent\textbf{When some background is helpful:} In some FSS cases, the background offers relevant context that is unavailable in the support mask, for accurate segmentation. However, many recent methods depend on support masks
to completely remove the background. 
We notice the person class frequently exists in the background of the support image in PASCAL-5$^i$~\cite{pascal}. 
Therefore,
in order to empirically validate the importance of leveraging relevant background information,
% relevance of background,
% in some challenging cases, 
we train VAT+MSI and VAT~\cite{VAT} by removing the segmentation mask of the person class when the person exists in the background of the support image. 
We compare
with or without the person class existing in the background of the support image in~Tab.~\ref{table:Ablation_study4} and~Fig.~\ref{fig:background_info}.
% This is because we notice the person class often exists in the background of the support image.
% We chose PASCAL-5$^i$~\cite{pascal} for this purpose as persons appear most frequently in the background.
After excluding the person class, the performance drop for VAT+MSI was much larger than VAT~\cite{VAT}. 

% This  validates that MSI particularly benefits from the common information (e.g., person) in the background between query and support images.

% Persons appear most frequently in the background in PASCAL.

% VAT+MSI shows a performance degradation when background information in the support set is unavailable. Compared to VAT+MSI, the degradation from VAT is very small because the background features are not utilized.

%Persons appear most frequently in the background in PASCAL. We remove persons in the background and found the performance drop of MSI was much larger than the baseline's. This validates that MSI particularly benefits from the common information (e.g., person) in the background between query and support images.

% We conjecture that this happens
% because 

% The training was conducted without person class existing in the background of the support image~(see Tab.~\ref{table:Ablation_study4}). MSI showed a performance degradation when background information in the support set was lost. This indicates that MSI uses background information in the support image to segment the target class accurately. 

% Notable point is surprisingly training without person in background still showed better performance than VAT~\cite{VAT}.

% We manually searched for the cases where background in a query image is similar to background in a support image and compared the performance the proposed method with the baseline VAT~\cite{VAT}~(see Fig.~\ref{fig:background_info}). 

\begin{figure}
  \centering
    \includegraphics[height=3.7cm]{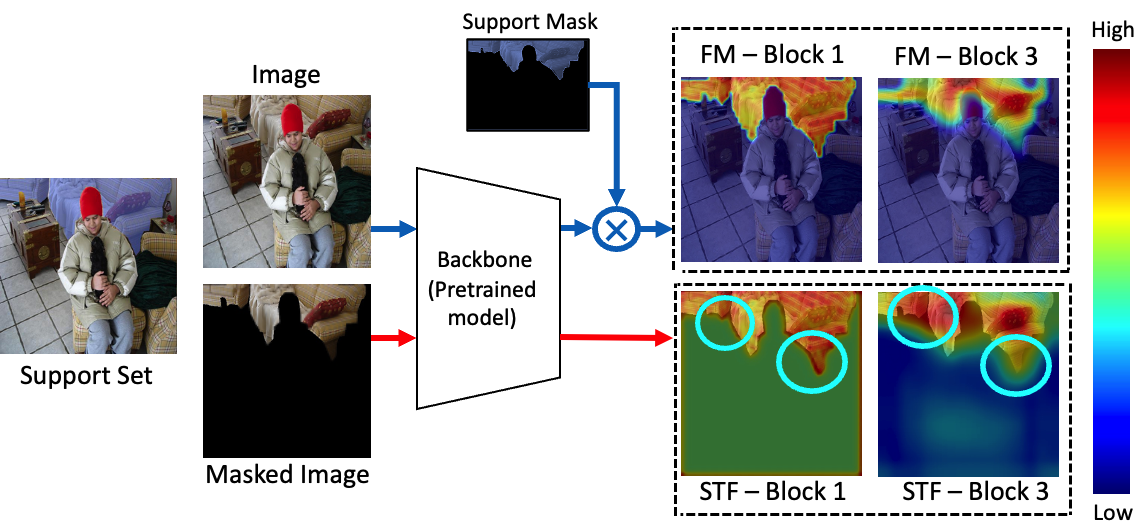}

   \caption{\small Feature map comparison between support target features~(STF) and feature masking~(FM). The cyan circles show that STF captures more fine-grained features near the target object boundary, however, FM struggles to capture the same. In deeper layers, the advantage of STF becomes more apparent (e.g. Block~3).} 
   %on COCO-20~$^i$~\cite{lin2015microsoft} when background information of support set and query image is similar to each other.}
   \label{fig:feature_analysis}
   \vspace{-3mm}
\end{figure}

\begin{figure}
  \centering
    \includegraphics[height=2.1cm]{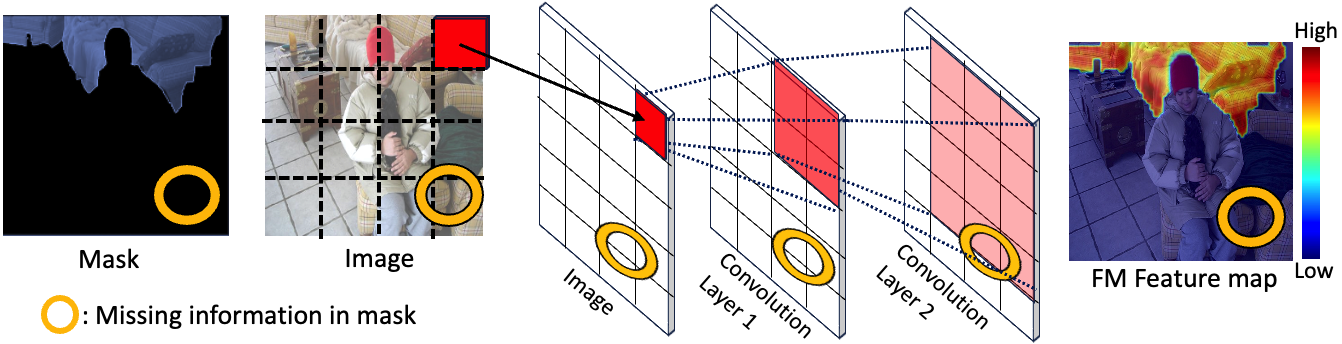}
   \caption{  
   \small Feature masking~(FM) overlooks the impact of the growing receptive field. Background deep features could contain target object information. FM thus loses the detailed target information by masking all features using a support mask. This becomes severe when the support mask is small/inaccurate.}
   
   \label{fig:receptive_field}
   \vspace{-3mm}
\end{figure}

\subsection{Ablation Study and Analysis}

\noindent\textbf{Why the masked support image is helpful:} (1)~Feature masking (FM) is utilized by several existing FSS models~\cite{PMM,PANet,PFENet,FWB,CANet,ASGNet,zhang2021fewshot,HSNet, VAT2,ASNet,BAM}, obtained by masking background features. The
target features in FM are not merely targeted object features. When extracting target features in FM, the features corresponding to background objects could also be present because of the enlarging receptive fields~\cite{receptive_analysis}. These background features interfere with obtaining target information. However, this interference can be eliminated by masking the background at the input image level in our MSI. (2)~STF holds more target features such as the texture and boundary of the target object than FM. In contrast, FM loses such information by masking features to remove the background (Fig.~\ref{fig:feature_analysis})~\cite{hmmasking}.

\noindent\textbf{How MSI helps for each case:} FM is
% the method 
utilized in most FSS works to remove the background 
% information 
in support features~\cite{PMM,PANet,PFENet,FWB,CANet}. (a) When the support mask fails to mask the same target class in a different location, unlike FM, SIF in MSI can still retain the target information (Fig.~\ref{fig:features}).
% because we obtain SIF by feeding the entire support image to a backbone without any masking. 
% The target information obtained with STF enables the encoder/decoder to selectively activate target features in SIF (Fig.~5). %~(Fig.~\ref{fig:receptive_field}). 
%
(b)~When the target class is minuscule, FM features retain minimal target information from masking downstream features~(Fig.~\ref{fig:receptive_field}). 
% by masking features with a very small mask.
However, in MSI, we mask support images at the input level.
% we extract target features from a masked support image. 
As such, we can retain fine-grained information in STF. Also, SIF could hold the target information missed by STF due to a small mask (Tab.~\ref{table:Ablation_study4}). 
% MSI not only utilizes STF but also exploits relevant background information in SIF~(Tab.~8).
%
(c) FM loses target boundary information~(Fig.~\ref{fig:boundary_error}) when removing background features with an inaccurate mask.
% FM performs feature masking to remove background features with an inaccurate mask, especially around boundaries. This results in losing target boundary information~(Fig.~6~\&~13sup.).
However, MSI can overcome this by exploiting additional target information in SIF and it facilitates a network in recovering fine-grained target details through the encoder/decoder.
%similar to (a) and (b)
(d) FM
% loses background information and 
cannot utilize the contextual features from the background.
% for helping locate the target. 
% Contextual features from background might have the information that allows the network to distinguish between the target and the background.
For example, if the support and query images have the same non-target object in the background, MSI uses $CS_1$  to learn contextual features
% to find similarity between SIF and QF 
and recognizes the same objects in QF and SIF,
and $CS_2$ to find the similarity between STF and QF to learn target-specific features to locate the target object (Fig.~\ref{fig:overall}, Eq.~\ref{eq:4}). Both $CS_1$ and $CS_2$ allow the network to recognize  non-targets~(Fig.~\ref{fig:background_info}, Tab.~\ref{table:Ablation_study4}) and leverage such information.

\noindent\textbf{Best features for SCM:} In Tab.~\ref{table:Ablation_study1}, we show results when using various combinations of features to obtain SCM. For this purpose, VAT~\cite{VAT} with ResNet50 backbone is used as a baseline and PASCAL-5$^i$ is chosen for training and testing the model. The ablation study reveals that harnessing both SIF and STF via concatenation achieves the best mIoU. %of 68.3\% among all others.
%
% Please see the supplementary text for visualizations of the distributions of the correlation values of SIF and STF.
Please see the supplementary text for visualizations of the correlation value distributions for SIF and STF.

\noindent\textbf{On different ways of fusing SIF and STF:}
% The proposed method, MSI, calculates cosine similarity between two features (STF and SIF) and QF, and concatenates the two generated correlation maps.
% The proposed method, MSI, calculates cosine similarity between two features STF, SIF, and QF, and concatenates the two generated correlation maps.
The proposed method, MSI, calculates cosine similarity between $\langle$STF, QF$\rangle$ and $\langle$SIF, QF$\rangle$, and concatenates the two generated correlation maps.
To see the effectiveness of this method, we experimented with different approaches to fuse SIF and STF to obtain SCM~(see Tab.~\ref{table:Ablation_study2}).
% Feature addition refers to adding features in element-wise before calculating cosine similarity to get correlation map.
Feature addition refers to the element-wise addition of features before calculating the cosine similarity to get the correlation map.
Correlation addition means simply adding two correlation maps element-wise. The attention method merges two correlation maps through 1x1 depth-wise attention based on Attention~U-Net~\cite{attention_unet}. Among all design choices for fusion, concatenating the two correlation maps showed the best performance. %The attention method also showed a greater improvement than baseline, VAT~\cite{VAT}. 

%When two features are merged, the number of parameters that a network needs to learn are reduced. Therefore, the attention-based method can also be recommended for the purpose of minimizing learning parameters.

% \subsection{Data augmentation.}
% In the VAT~\cite{VAT2}, data augmentation was not recommended to use because it lowered the performance. However, when using our proposed MSI method, VAT~+~MSI learned more stably and showed better performance when using cat data augmentation. Unlike the existing method, by giving strong target information of STF, rather than picking up noise in the input data through data augmentation to confuse the network, it shows that it is useful for learning more diverse input data and shows a robust training network.

\begin{figure}[t]
  \centering
    \includegraphics[height=3.3cm]{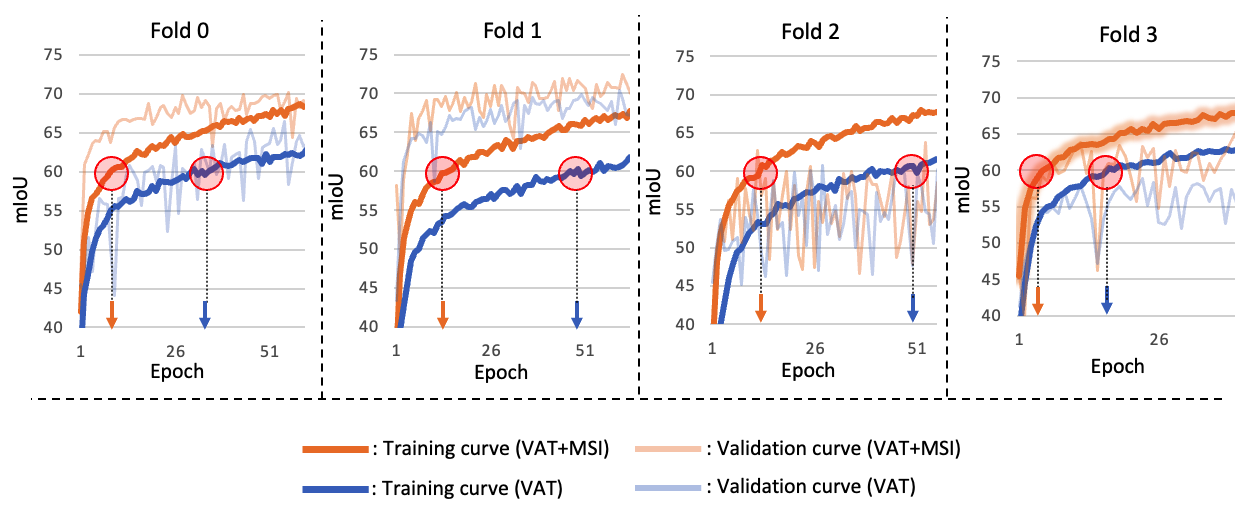}

   \caption{Train. and val. profiles of VAT~\cite{VAT} and VAT~+~MSI on PASCAL-5$^i$ with ResNet50. VAT~+~MSI provides 3.3x faster convergence (to reach 60\% in mIoU) on average than VAT. Red circles indicate when the training accuracy reaches 60\% in mIoU.}
   \label{fig:train_curve}
\end{figure}

%\subsection{Fast training}
\noindent\textbf{Fast training convergence:} We notice that MSI with VAT~\cite{VAT} and ResNet50 on PASCAL-5$^i$, VAT~+~MSI, provides 3.3x faster convergence (i.e.~to reach mIoU of 60\%) on average than VAT~(Fig.~\ref{fig:train_curve}), along with a remarkable improvement in performance. Similarly with HSNet, HSNet+MSI allows faster convergence  by 4.5x on average. See supplementary for plots.

\section{Conclusion}

We proposed a new method, MSI, maximizing the information of the support set, deviating from the masking method~\cite{Zhang2020SGOneSG} widely used in many FSS models~\cite{PMM,PANet,PFENet,FWB,CANet,ASGNet,zhang2021fewshot,HSNet, VAT2,ASNet,BAM}. The joint exploitation of SIF and STF develops a synergy between them, which allows handling several challenging FSS cases. The results show significant performance improvements across all benchmarks on recent, strong FSS baselines~(HSNet~\cite{HSNet}, ASNet~\cite{ASNet}, and VAT~\cite{VAT}). 

%SIF holds the entire support image features and STF has detailed target information. Their complementary nature  in a significant performance improvements across all benchmarks with recent FSS baselines~(HSNet~\cite{HSNet}, ASNet~\cite{ASNet}, and VAT~\cite{VAT2}). Also, the usefulness of the method is demonstrated through case studies.

% \section{contents to add Supplement material}
% 1. Training profile figure (COCO, FSS with HSNet, VAT)

% 2. attention based merge architecture

% 3. more visual comparison figures

% 4. failure cases sutdy

% 5. Data augmentation explanation

% 6. Why ASNet performance is inferior to others

% 7. boundary analysis 

\section*{Acknowledgment}
This work is supported in part by NSF Grant \#1955404 and \#1955365.
%\hl{this section maybe placed in page 9}

%
% \section{Story to add}

% Main story: Feature masking is used frequently for the most of few-shot segmentation works. This work found that input masking and utilizing entire support image to get whole features improves all the very recent baselines. Input masking features work as attention map. Therefore we don't require feature masking anymore. Proposed appraoch with VAT res101 performed SOTA on PASCAl-5. Proposed method is the first work reaching the 70 in mIoU among all the other works. The impressive performance improvements observed in across all benchmarks. Especially, with VAT D

% 1.3 show that difference when we fully utilize all support images with fm masking features

% 1.4 show that input masking works as attention 

%%%%%%%%% REFERENCES
{\small
\bibliographystyle{ieee_fullname}
\bibliography{iccv/main}
}

%%%% Appendix -- comment these two lines for main manuscript submission
% \clearpage
% \include{iccv/appendix.tex}

\end{document}